1

# Deep Learning Provides Rapid Screen for Breast Cancer Metastasis with Sentinel Lymph Nodes

Kareem Allam[1], Xiaohong Iris Wang[1], Songlin Zhang[1], Jianmin Ding[1], Kevin Chiu[2], Karan Saluja[1], Amer Wahed[1], Hongxia Sun[1], Andy N.D. Nguyen[1]

[1]Department of Pathology and Laboratory Medicine
[2]Medical School
 University of Texas Health Science Center-Houston, Medical School, Texas 77030
**ABSTRACT**
Deep learning has been shown to be useful to detect breast cancer metastases by analyzing whole slide images (WSI) of sentinel lymph nodes (SLNs); however, it requires extensive scanning and analysis of all the lymph node slides for each case. Our deep learning study focuses on breast cancer screening with only a small set of image patches from any SLN to detect changes in tumor environment and not in the tumor itself. This study involves breast pathologists in our department and uses our in-house breast cancer cases and WSI scanners. We design a convolutional neural network in the Python language to build a diagnostic model for four diagnostic categories (macrometastasis, micrometastasis, isolated tumor cells, and negative metastasis). SLNs with macrometastasis and micrometastasis are defined as positive cases; while those with isolated tumor cells only or true negative for metastatic tumor cells are defined as negative cases. We obtained WSIs of Hematoxylin and Eosin-stained slides from 34 cases with near equal distribution in 4 diagnostic categories. A total of 2720 image patches, from which 2160 (79%) were used for training, 240 (9%) for validation, and 320 (12%) for testing. Interobserver variation was also examined among 3 users. The test results showed excellent diagnostic results: accuracy (91.15%), sensitivity (77.92%), specificity (92.09%), positive predictive value (90.86%), and negative predictive value (80.66%). No significant variation in results was observed among the 3 observers. This preliminary study provided a proof of concept for incorporating automated metastatic screen into the digital pathology workflow to augment the pathologists' productivity. Our approach is unique since it provides a very rapid screen rather than an exhaustive search for tumor in all fields of all sentinel lymph nodes.

**Corresponding Author:**
Andy N.D. Nguyen, MD, MS
Department of Pathology and Laboratory Medicine
University of Texas Health Science Center-Houston, Medical School
6431 Fannin Street MSB 2.292
Houston, Texas 77030
Telephone: (713) 500-5337
Fax: (713) 500-0712
Email: Nghia.D.Nguyen@uth.tmc.edu

**INTRODUCTION**
In surgery for a patient with breast cancer, the surgeon finds and removes the first lymph node(s) to which a tumor is likely to spread (called SLNs). To do this, the surgeon injects a radioactive substance and/or a blue dye into the tumor, the area around it, or the area around the nipple. Lymphatic vessels will carry these substances along the same path that the cancer would take. The first lymph nodes that dye or radioactive substance travels to are the SLNs. The evaluation of breast SLNs is an important component of treatment. Patients with a SLN positive for metastatic cancer will receive a more aggressive clinical management, including axillary lymph nodes dissection.

The manual microscopic examination of SLNs is time-consuming and laborious, particularly in cases where the lymph nodes are negative for cancer or contain only small foci of metastatic cancer [1]. SLNs can be grouped into two types: positive for metastasis or negative for metastasis. Of the ones that are positive for metastasis, they can have macrometastasis (tumor region of at least 2.0 mm) or micrometastasis (tumor region of at least 200 cells or with size between 0.2 mm and 2.0 mm). Of the ones that are negative, they can be truly negative, or have isolated tumor cells (ITC) only, which is a tumor region of up to 200 cells and/or smaller than 0.2 mm [2]. Potential morphologic features for metastasis include pleomorphic nuclei and some features of the tumor microenvironment, including lymphocytic infiltrates in the stroma, the sinus, and follicular hyperplasia [3].

Due to the large number of SLNs to screen for breast cancer metastasis, histopathologic screening often presents a challenge to the pathologists. An automated diagnosis using digital images would be helpful to assist the pathologist in daily work. In this study, we investigate how automated screening methods can be combined with microscopic examination by pathologists to achieve better accuracy. We focus on using reactive morphology in non-tumor areas of lymph nodes to predict positive metastasis. To analyze slides by automated techniques, it is first necessary to scan the slide into the computer's data storage. This process is called whole slide imaging (WSI). Techniques of WSI involve scanning and compressing the images before they are analyzed [4]. WSI offers many advantages such as ease of slide sharing and image analysis [5].

Previous attempts to digitally classify histologic images were based on specific criteria (such as nuclear shape, nuclear size, texture, etc.) [3,6]. They turned out not to be successful [7]. Attention has turned to machine learning. Machine learning can be defined as software algorithms that can learn from and make predictions on data. This gives the software the ability to learn without being explicitly programmed. There are numerous machine learning methods. Some examples are decision trees, cluster analysis, support vector machines, random forests, Bayesian networks, regression analysis, and neural networks [7]. Neural networks consist of multiple artificial nodes ("neurons") connected to form a network for prediction/classification [8]. This is inspired by biological neural networks. Early generations of neural networks used supervised training, but this has some disadvantages. One disadvantage is that the parameters (such as the strengths of the connections between the neurons) may not converge, leaving no solution. Another disadvantage is that it may not scale well.

Deep learning is the most recent and most disruptive method of machine learning; it is based on neural networks. In 2006, major breakthroughs in deep learning started. One is unsupervised learning, which allows a network to be fed with raw data (no known outcomes) and discover the representations needed for detection or classification. Another is the use of multiple layers in the network, which allows it to extract high-level and complex data representations and avoid some of the problems of older neural networks. Since such methods perform many operations in parallel, they can be speeded up by using graphics processing units (GPUs). Studies have been done to assess the reproducibility of deep learning algorithms by using them to identify the tissue of origin from 512x512 pixel tiles. The performance of the algorithm was better than pathologists viewing the same tiles [9]. Deep learning techniques, especially third generation neural networks called convolutional neural networks (CNN or ConvNet), have quickly become the state of the art in computer vision [10]. The ventral visual pathway is organized as a hierarchical series of four interconnected visual areas called Brodmann areas. Neurons in early areas, such as area V1, respond to comparatively simple visual features of the retinal image, while later areas, such as area V4, respond to increasingly complex visual features. The specialization of receptor cells is incorporated into the design of the CNN as pairs of convolution operators followed by a pooling layer (Figure 1) [11]. Convolution is an operation in image processing using filters to modify or detect certain characteristics of an image (such as Smooth, Sharpen, Intensify, Enhance). In CNNs, it is used to extract features of images. Mathematically, a convolution is done by multiplying the pixels' values in an image patch by a filter (kernel) matrix and then adding them up (Figure 2). This operation is also called a "dot product". By moving the filter across the input image, one obtains the final output as a modified filtered image. The CNN consists of interleaved convolutional and max pooling layers and then a final fully connected layer (Figure 1) [12]. The convolutional layers (C) perform 'feature extraction' consecutively from the image patch to higher level features. The max pooling layers (S) reduce image size by subsampling. The last 'fully connected' layers (F) provide prediction. Convolutional neural networks have been used to generate heat maps of tumors and tumor-infiltrating lymphocytes [13, 14]. Big companies are analyzing large volumes of data for business analysis and decisions, using deep learning technology (Google's search engine, Google Photo, automobile companies with self-driving cars, and IBM's Watson).

The application of deep learning to digital pathology imaging has a promising start; it could impact personalized diagnostics and treatment. Deep learning has also been considered in interpreting and integrating multiple sources of information in pathology (histology, molecular, etc.) [15]. Recent studies have shown promising results in using deep learning to detect breast cancer in whole slide imaging of SLNs (examples: Camelyon16, ICIAR 2018)[16, 17]. However, they require extensive scanning and analysis of all the lymph node slides for each case. We explore how deep learning could be used for breast cancer screening with only a small set of image patches (5 patches) from any SLN. Our goal is to detect changes in the tumor microenvironment and not the tumor itself (Figure 3). Our approach is unique since it provides a very rapid screen rather than an exhaustive search for tumors in all fields of all lymph nodes. We also set out to examine the feasibility of looking at either negative or positive slides (in the uninvolved area) to predict metastasis. The tumor microenvironment has been shown to be important in diagnosing the tumor [18]. We examined three areas of interest: interfollicular lymphocyte-rich area, follicles, and the sinus (Fig. 3) to see which is best for predicting metastasis. Previous studies have examined the tumor-infiltrating lymphocytes [19, 20].

Interobserver variation was also examined among different users. We assessed variation in predictive results with data obtained by 3 users.

**Fig. 1** The CNN deep learning model

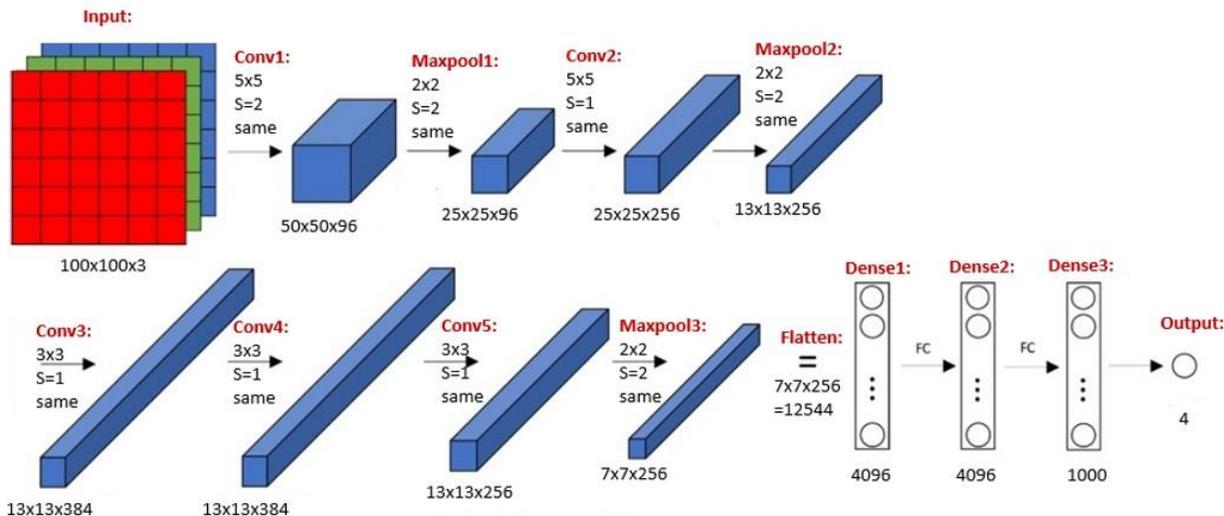

**Fig. 2** Convolution of input with kernel

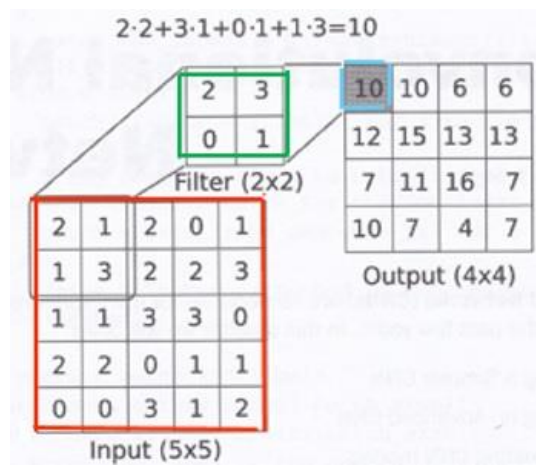

**MATERIALS AND METHODS**
Our study was approved by the Institution Review Board at the University of Texas Health Science Center. We obtained WSIs of SLNs using Motic scanners (Motic Easy Scan, Motic Instrument, Richmond, BC, Canada) in the pathology department of the University of Texas-Houston Medical School to obtain our data. The Motic Digital Slide Assistant software (by the same company) was used to view the WSIs. SnagIt (TechSmith Corp, Okemos, Michigan, USA) was used to capture and automatically save image patches (100x100 pixels) in files with JPEG format. Our study includes 34 cases with near equal distribution in 4 diagnostic categories:
1. Macrometastasis: 10 cases
2. Micrometastasis: 8 cases
3. Isolated tumor cells (ITC): 6 cases
4. Negative: 10 cases

A positive WSI and a negative WSI were selected from each positive case, and two negative WSIs were selected from each negative case to obtain a total of 68 WSIs. For each WSI, 40 image patches were obtained for a total of 68x40 = 2720 image patches. Of the 2720 image patches, 2160 images (79%) were used for training the model, 240 (9%) were used for validation, and 320 (12%) for testing. We designed the CNN model using the Python language together with the TensorFlow and Keras libraries. The model ran on 64-bit Windows 10 Professional edition. Keras allows for parallel computing using graphics processing units (GPUs) with the Compute Unified Device Architecture (CUDA) by NVIDIA (Santa Clara, CA, USA). The hardware was 9th Gen Intel® Core™ i7 9700 (8-Core, 12MB Cache, 4.7GHz), 32GB RAM (DDR4 at 2666MHz), and NVIDIA® GPU (GeForce RTX™ 2070, 8GB GDDR6, 2304 CUDA cores). Our deep learning model used 14 layers, including convolution, max-pooling, and dense layers (Figure 1).

**RESULTS**

We looked at different areas of interest in WSIs to see which one would be of most predictive value (positive vs negative metastasis): 1. Interfollicular lymphocyte-rich areas, 2. Follicles, and 3. The sinus (Figure 3). The preliminary results indicated that areas containing interfollicular lymphocytes are of most predictive value (full results not reported in this article). Subsequently, our study has been focusing on this parameter alone.

The image-by-image accuracy of user 1 was found to be 161/320 = 50.31% (Table 1). When the diagnoses were grouped by rank (i.e., diagnoses 0 and 1 are considered negative, 2 and 3 are considered positive), significantly better accuracy was achieved at 275/320 = 85.93% (Table 2). For each test case, the predicted diagnosis was combined from the prediction for 5 images (at least 3 or more must agree), a process known as "majority voting" (see examples in Table 3). This led to a higher accuracy of 92.18% (59 sets/64 sets). When majority voting was used, we obtained the data in Table 4.

From this, we calculated the accuracy, sensitivity, specificity, positive predictive value (PPV), and negative predictive value (NPV) for user 1:
Accuracy = 59/64 = 92.18%
Sensitivity=27/(27+5)=84.4%
Specificity=32/(32+0)=100%
PPV=27/(27+0)=100%
NPV=32/(32+5)=86.5%

In a similar manner, we calculated those values for the other 2 users. The results were tabulated in Table 5. No significant variation was observed among the 3 observers. The average results were as following: Accuracy = 91.15%, Sensitivity=77.92%, Specificity=92.09%, PPV=90.86%, NPV=80.66%

**Fig. 3** Areas of interest in WSI

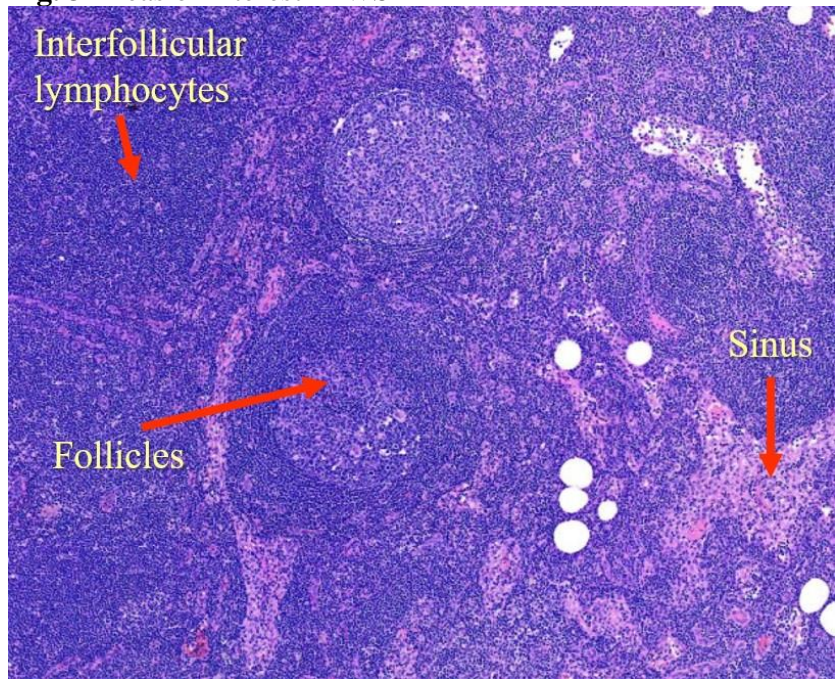

Table 1. Image-by-image accuracy

| | | Predicted Diagnosis | | | |
|---|---|---|---|---|---|
| **Observed Diagnosis** | | Negative (0) | ITC (1) | Micro Met (2) | Macro Met (3) |
| | Negative (0) | **42** | 36 | 2 | 0 |
| | ITC (1) | 20 | **54** | 3 | 3 |
| | Micro Met (2) | 12 | 2 | **65** | 1 |
| | Macro Met (3) | 22 | 1 | 57 | **0** |

Accuracy: 161/320 = 50.31%

Table 2. Grouped ranking

| **Predicted Diagnosis** | | | |
|---|---|---|---|
| **Observed Diagnosis** | | Negative (0) or ITC (1) | Micro Met (2) or Macro Met (3) |
| | Negative (0) or ITC (1) | 152 | 8 |
| | Micro Met (2) or Macro Met (3) | 37 | 123 |

Accuracy: 275/320 = 85.93%

Table 3. Examples of majority voting process

| Result for each image | Case-by-case (set of 5) |
|---|---|
| Observed dx=3, Predicted dx=1<br>Observed dx=3, Predicted dx=0<br>**Observed dx=3, Predicted dx=2**<br>Observed dx=3, Predicted dx=0<br>**Observed dx=3, Predicted dx=2** | 2/5 → Incorrect |
| **Observed dx=2, Predicted dx=2**<br>**Observed dx=2, Predicted dx=2**<br>Observed dx=2, Predicted dx=0<br>Observed dx=2, Predicted dx=0<br>**Observed dx=2, Predicted dx=2** | 3/5 → Correct |
| **Observed dx=1, Predicted dx=1**<br>**Observed dx=1, Predicted dx=1**<br>**Observed dx=1, Predicted dx=1**<br>**Observed dx=1, Predicted dx=0**<br>**Observed dx=1, Predicted dx=0** | 5/5 → Correct |
| **Observed dx=0, Predicted dx=1**<br>Observed dx=0, Predicted dx=2<br>**Observed dx=0, Predicted dx=0**<br>**Observed dx=0, Predicted dx=1**<br>**Observed dx=0, Predicted dx=1** | 4/5 → Correct |

Table 4. Data for majority voting with group ranking for user 1.

| | Predicted negative | Predicted positive |
|---|---|---|
| Observed negative | 32 | 0 |
| Observed positive | 5 | 27 |

Accuracy: 59/64 = 92.18%

Table 5. Accuracy, sensitivity, specificity, PPV, and NPV for all 3 users.

| User | Accuracy | Sensitivity | Specificity | PPV | NPV |
|---|---|---|---|---|---|
| User 1 | 92.19 | 76.88 | 95 | 93.89 | 80.42 |
| User 2 | 87.5 | 78.75 | 89.38 | 88.11 | 80.79 |
| User 3 | 93.75 | 78.12 | 91.88 | 90.58 | 80.77 |
| **Means** | **91.15** | **77.92** | **92.09** | **90.86** | **80.66** |

## DISCUSSION

Deep learning has been shown to be useful for the identification of breast cancer metastases by analyzing whole sections of slide images of SLNs [12, 21]. Our study focuses on breast cancer screening using deep learning with only a small set of image patches from any SLN (positive or negative) to detect changes in the tumor microenvironment and not the tumor itself. Our approach is unique since it provides a very rapid screen rather than an exhaustive search for tumors in all fields of all lymph nodes. We obtain excellent predictive results for cancer metastasis in this study, which provide a proof of concept for incorporating automated breast cancer metastatic screen into the digital pathology workflow to potentially augment the pathologists' productivity. This could have a significant impact on health economics.

Some limitations of this study are:
1. The model was only validated on one hardware platform (Motic scanner),
2. Representative images require preselection of lymphocyte-rich areas,
3. Lack of explicit diagnostic criteria (inherent to deep learning).

Our preliminary study nevertheless provided a proof of concept for incorporating automated breast cancer screen using digital microscopic images into the pathology workflow to augment the pathologists' QA. Future studies will need to (a) include more hardware platforms and many more cases for training and validation, and (b) use automated segmentation of WSIs for lymphocyte-rich areas.

## CONCLUSION

We obtained excellent predictive results for cancer metastasis from this study: 91% accuracy, 78% sensitivity, and 92% specificity using a set of 5 random image patches (100x100 pixels) from each test case. There is a potential role for this model in clinical work as a QA tool. If a case is positive by histology, a final diagnosis of metastasis can readily be made. For cases that are negative by histology, our model can be used to screen for metastasis. If the screen is negative, a final diagnosis of negative metastasis can be made, and if the screen is positive, the slide can be re-examined to either find the metastases or to make a final diagnosis of negative metastasis if none is found. In this way, the model serves as an extra checking step to help detect metastases that otherwise would be missed with just manual examination.